# Automatic Language Identification System for Hindi and Magahi

Priya Rani, Atul Kr. Ojha, Girish Nath Jha
JNU, New Delhi
{pranijnu, shashwatup9k, girishjha}@gmail.com

## Abstract

Language identification has become a prerequisite for all kinds of automated text processing systems. In this paper, we present a rule-based language identifier tool for two closely related Indo-Aryan languages: Hindi and Magahi. This system has currently achieved an accuracy of approx 86.34%. We hope to improve this in the future. Automatic identification of languages will be significant in the accuracy of output of Web Crawlers.

## 1. Introduction

Code-mixing is a common phenomenon in countries like India where five different language families co-exist. According to a report issued by Microsoft Research, 95% of the languages used by Indians are mixed (Chittaranjan, 2014). This paper focuses on two very closely related Indo-Aryan languages: Hindi and Magahi. Hindi being a scheduled/official language (languages which are included in the 8th schedule of constitution of India.), is used for official purpose, spoken in north, western, central and eastern parts of India. Whereas, Magahi is a non-scheduled or non-official language spoken in eastern states of India including Patna, Gaya, Jehanabad, Munger, Begusarai, Hazaribagh, Nalanda districts of Bihar, Ranchi district in Jharkhand, some parts of Orissa and Malda district in West Bengal (Kumar, 2011). Even though due non-experts consider Magahi as one of the dialects of Hindi, linguists understand it as a separate language owing to significant difference between both the languages. According to Census 2001, Hindi is spoken by 534,271,550 people and Magahi speakers count up to 14,046,400.[1] In this paper, we report a rule-based language identifier tool for Hindi and Magahi. The immediate goal is to identify the language of a given text. The paper demonstrates the function, experimental set-up, efficiency and limitations of the tool.

### 1.1 Motivation of the Study

Language Identification is the process of finding the natural language in which the content of the text is encoded. (Garg ét al., 2014). It is an extensive research area used in various fields such as machine translation, information retrieval, summarization etc. It is easier to distinguish two languages belonging to different language families, and with different typological distributions. It becomes even more easier to distinguish two languages if they are encoded in different scripts. However, the identification task becomes challenging when the two languages belong to the same language family and share many typological and areal features. In this paper, we will develop a tool to identify two closely related languages Hindi and Magahi only that share many typological and areal features and belong to the same language family. Despite these relatedness, these languages differ from each other in many respect. We will focus on those differences and use them to develop the tool.

### 1.2 Features of Hindi and Magahi

The section deals with some basic linguistic features in an attempt to differentiate between Hindi and Magahi.

(a) A primary difference between both the languages is that while Magahi is a nominative-accusative language, Hindi is an ergative language.
   For example:

Magahi
     rəm-mɑ  sit̪-wa   ke əm-mɑ
     ram-PRT sita-PRT  to mango-PRT
     delkai
     givePST.
     Translation- "Ram gave mango to Sita"

Hindi
     ram-ne  ʃit̪a-ko am    dija
     ram-ERG Sita-DAT mango  give-PST

---

[1] https://www.ethnologue.com/language/hin and https://www.ethnologue.com/language/mag

Translation- "ram gave mango to sita."

(b) Magahi, like other eastern Indo-Aryan languages and unlike Hindi do not show number and gender agreement. It reflects agreement with person and honorificity. Whereas Hindi shows agreement with phi features i.e person , number and gender as well as posses honorific agreement.
For example:

Magahi
 i. sit-wa ɟa he
    sita-PRT go AUX-3P.NH
    "Sita is going."
 ii. apne ɟait hatʰin
    You.H go AUX-H
    "you are going"
 iii. həmni ɟa hi
    we.NH go AUX-2P.NH
    "we are going"

Hindi
 iv. ʃiʈa ɟa rahi hai
    sita go PROG.F AUX.3SG
    "Sita is going"
 v. aap ɟa rahe hain
    You go PROG.H AUX.2SG.H
    "you are going"
 vi. həm log ɟa rahe hain
    we all go PROG AUX.1PL
    "we are going."

(c) Numeral classifiers are prominent in Magahi but Hindi lacks them. For example:

| Hindi | ek | ɖo | tɪnə |
|---|---|---|---|
| Magahi | e-**go** | ɖu-**go** | tɪn-**go** |
| Translation | one | two | three |

(d) Nouns have two basic forms in Magahi : Base form and Inflected form. The particles -wa, -ia, -ma, -a are added to the base form to construct an inflected form. The nominal particles -ia, -a, -ma and -a are allomorphs of base form -wa. (Alok, 2010). These are used to show different linguistic features. These particles are adddded to proper names as well. Whereas nouns in Hindi have only one form.
For example:
Magahi

|     | Form1 | Form2 |
|---|---|---|
| i. | gʰər | gʰər-wɑ |
|    | house | house-PRT |
| ii. | ɑm | əm-mɑ |
|    | mango | mango-PRT |
| iii. | rɑm | rəm-mɑ |
|    | Ram | Ram |

Hindi
 iv. gʰər
    house
 v. ɑm
    mango

(e) Verbs shows some interesting and complex features in both languages. The difference lies in inflections that they take. Magahi present tense is unmarked, past tense is marked with '-l-` and future with '-b-`. In hindi the past markers are '-a`,'-j-`,'-i` and future marke is the optative marker '-ga`.
For example:
Magahi
 i. ʊ sʊt-l-o
    he sleep-PST-NH
    "he slept"
 ii. tʊ sʊt-b-ə
    you sleep-FUT-2P-NH
    "you saw"

Hindi
 iii. tum -ne dekʰ –a
    you-ERG see-PST.M.SG
    "you saw"
 iv. tu dekʰe-ga
    you see-OPT.FUT.M.2SG
    "You will see"

(f) In Magahi a plural marker '-ən` is added to form plural constructions but this marker is absent in numeral constructions. Whereas in Hindi, plural constructions are formed by adding nasalisation irrespective of any form of constuction. For example:

|  | Singular | Plural |
|---|---|---|
| Magahi | ləikɑ | ləik-ən |
|  | boy | boys |
|  | e-go ləikɑ | ɖu-go ləikɑ |
|  | one-CLF boy | two-CLF boy |
| Hindi | ləɽka | ləɽke˜ |
|  | boy | boys |
|  | ek ləɽka | do ləɽke˜ |
|  | one boy | two boys |

(g) Hindi and Magahi both differ in their lexicon as well.
For example:

| Hindi | si:r | ɖʰoop |
|---|---|---|
| Magahi | matʰa | rauɖa |
| Translation | head | sunrays. |

(h) Adjectives, like nouns, also have two forms in Magahi: a base form and an inflected form. The inflected nouns always take inflected adjectives. Concord between an adjective and a noun is inflected with number, gender (it should be noted that concord inflecting gender has to be natural sex in case of animates and not the Noun class as it is used in Hindi) and also familiarity (Alok, 2010). Hindi adjectives too show inflection but concord is only with number and gender (both natural and grammatical).

For example:

Magahi
  i. kəri-kɑ           ləik-wɑ
     black-SUF-M       boy-PRT
     "the black boy"
  ii. kəri-k-iː        ləiki-ɑ
     balck-SUF-F       girl-PRT
     "the black girl"
  iii. kəri-k-ən       ləik-w-ən
     black-SUF-PL      boy-PRT-PL
     "the black boys"

Hindi
  iv. kala             ləɽka
     black             boy
     "Black boy"
  v. kali              ləɽki
     black             girl
     "black girl"
  vi. kalə             ləɽkə
     black             boys
     "black boys"

## 2. Literature Review

This section outlines a brief literature survey of Currently, no tool exists that can identify Magahi from Hindi. One of the reasons for this gap is that Magahi is a less-resourced language. There is a significant lack of computational resources in this language where one can find only a Magahi POS tagger, Magahi monolingual corpus, and Magahi Morph Analyser available (Kumar et al., 2011; Kumar et al., 2012; and Kumar et al., 2016). Several language identification tools have been developed in Indian languages such as (a) In 2008, OCR-based Language Identification tool was developed by Padma and Vijaya which gave 99% accuracy (Padma et al., 2008). (b) In 2014, text-based language identification system were developed for Devanagari script ( Indhuja et al., 2014). (c) In 2016, researhers developed a language identifier system for under-resourced languages and it was based on lexicon algorithm which gave an accuracy of 93% (Selamat, 2016).

(d) And, in 2017, Patro and others developed language identification tool to disinguish between English and Hindi text based on likelines estimate method with an accuracy of 88%. In this experiment they used social media corpus (Patro et al., 2017).

## 3. Experimental Set-up

This section isdivided into four sub-sections. It talks about corpus collection and creation, lexicon data-base, extraction of the suffixes, and architecture of the language identifier.

### 3.1 Data Collection

We have colletcted Magahi and Hindi corpora of 19,884 and 2,00,000 sentences respectively. Magahi data has been taken from the website https://github.com/kmi-linguistics/magahi (Kumar et al., 2016) and Hindi has been crawled from news and blog websites such as Amar Ujala, Live Hindustan, Dainik Jagran, Dainik Bhaskar etc.. We have also used Hindi monolingual corpus from WMT shared task (Bojar et al. 2014) and Indian Language Corpora Initiative (Jha 2010, and Bansal et al. 2013)

### 3.2 Creation of Lexicon database for Magahi and Hindi

The creation of lexicon database has been prepared using two approaches:

**(a)Prepration of unique words:**

| Magahi | | Hindi | |
|---|---|---|---|
| Word | Frequency | Word | Frequency |
| आउ | 5097 | हस | 19580 |
| ऊ | 2857 | कर | 19141 |
| गेल | 2162 | किया | 12839 |
| ओकरा | 1443 | गया | 11602 |
| हलइ | 1119 | अपने | 10361 |
| कहलक | 951 | रहे | 9303 |
| देलक | 713 | दिया | 8263 |

Table1: Frequency of Unique words from Magah and Hindi

The unique words for each of the language were extracted using ILDictionary[2], a java-based tool used to create frequency database. The unique words database consisted of 28,548 tokens for Magahi and 1,20,262 tokens for Hindi. In Table1, some words with their frequencies are given.

**(b) Extraction of multiple word dictionary:**

| Magahi | Hindi |
|---|---|
| हमरा जरूर | पहुंच गया |
| अज्ञथ हथ्न | ठेका मजदूरों के |
| कलेजा काढ़ के | बैठकर खाने का |

Table 2: Example of Multiple word dictionary

The multiple word groups were prepared upto trie-gram extracted from the corpora. And, for Magahi, we have also included a Morphological Analyser dictionary[3]. Some examples are presented in the table below.

### 3.3 Extraction of Suffixes

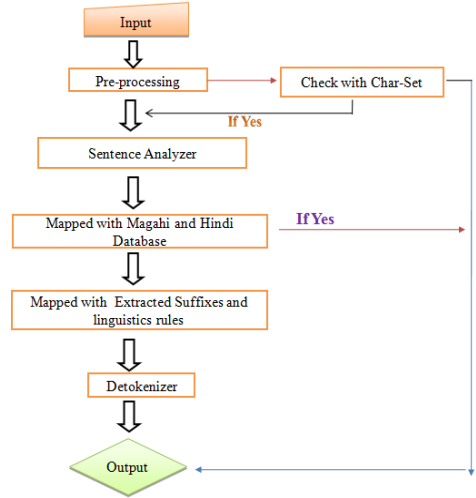

Figure 1: Extracted suffixes of Magahi and Hindi

Suffixes (index) up to 3 characters were extracted from both corpora. Total number of extracted unique suffixes are 8,715 in Magahi

---

2 http://sanskrit.jnu.ac.in

3 www.kmiagra.org/magahi-morph

---

and 8,629 in Hindi. Approximately 38.63% of suffixes in these langauges are same such as हित, हलक, ऑलेथा, धक, तैय, डा etc.

### 3.4 Architecture of Language Identifier

The figure demonstrated below presents the system architecture of the Language Identifier.

Figure 2: Architecture of Language Identifier

When a user inputs text to the tool, it first goes to the pre-processing section. This section is mapped with the Devanagri char-set. If the input text is in Devanagari then it is sent to sentence analyzer, else it goes directly to the output where tool displays the text belongs from another language. During pre-processing, if some tokens exist in other script then a hidden value is given to those tokens. In the next step, the input text goes to the sentence analyzer where it is tokenized at the word level. After tokenization, it goes for mapping with Magahi and Hindi lexicon data-base simultaneously. If text (tokens or combination of tokens) is mapped with Magahi lexicon data-base then the output "The text is Magahi" is displayed. If the text is mapped with Hindi database then the output "The text is Hindi" is displayed. When the text does/does not matches with both langauges then the system extracts suffix of each word of upto 3 characters. The extraced suffixes are first mapped with Magahi suffixes, through a file containing linguistic rules. If the rule and suffixes do not follow each other then the system cheks Hindi suffixes and its linguistic rules. Thereafter an output is generated in accordance with the mapped lingustic rules. Else an output "Text is of other language" is generated. Before generating the final output, the tokens are detokenized. The lingustic rules were prepared on the basis of distinguishing lingustics features of Magahi and Hindi and on the basis of their respective lexicon data-base. The current working system follows the rules on the basis of section 1.2 lingustic features only.

## 4. Evaluation and Analysis

This system has been evaluated on 2,000 sentences. These sentences came from Hindi, Magahi and other languages. The accuracy of the system has bee evaluated as 86.34%.

The system encountered an error rate of 13.66%. Magahi being a substratum language and Hindi being a superstratum, many lexical items are borrowed in Magahi from Hindi, such as - "फाइल नई दिल्ली ।". The borrwed words create problem the classification of languages. During system analysis, we found other major issues - the system's inability to distinguish between the Magahi and Hindi Named Enitities and spelling/typo errors. The system did not prove effective in its ability to tackle short sentences etc. which reduced the system accuracy. Examples of these issues are presented below:

(a) का हो रामौतार ।
(b) तू कौन हें/हे ।
(c) मात्र पचास रूपइया।
(d) उज्जर बाल ।

(a) type of examples have been identified for both langauges and error takes place due to the presence of named entity.
(b) is a Hindi sentence which has a typo/spelling error which resulted in a structure similar to Magahi.
(c) and (d) type of sentences can appear in both languages. Such short sentences (upto three words) contain words which are common in both languges. However the system identified these as Hindi instead of Magahi.

## 5. Conclusion

In this paper, we have presented a rule-based language identifier tool to identify a less-resourced language, Magahi, from Hindi. Magahi being closely related to Hindi and a substratum of the, pose greater challanges than unrelated languages.

Future work consists of fixing the above mentioned errors and increasing accuracy of the system. We believe writing heuristics verb anlysis rule can bring significant improvements in the system. We also plan to plug this tool with ILCralwer to improve crawling accuracy. The ILCrawler is used to create the computational framework for collecting Magahi corpus.

## 6. Acknowledgment

We would like to thank Dr. Ritesh Kumar and his team for providing the Magahi corpora. We would also like to acknowledge the efforts of our colleagues Deepak, Akanksha and reviewers for providing their valubale inputs to improve paper quality.